# Real time text localization for Indoor Mobile Robot Navigation

Kazem Ghazanfari, Saeed Shiri

*Abstract*— Scene text is an important feature to be extracted, especially in vision-based mobile robot navigation as many potential landmarks such as nameplates and information signs contain text. In this paper, a novel two-step text localization method for Indoor Mobile Robot Navigation is introduced. This method is based on morphological operators and machine learning techniques and can be used in real time environments. The proposed method has two steps. At First, a new set of morphological operators is applied with a particular sequence to extract high contrast areas that have high probability of text existence. Using of morphological operators has many advantages such as: high computation speed, being invariant to several geometrical transformations like translation, rotations, and scaling, and being able to extract all areas containing text. After extracting text candidate regions, a set of nine features are extracted for accurate detection and deletion of the regions that don't have text. These features are descriptors for texture properties and are computed in real time. Then, we use a SVM classifier to detect the existence of text in the region. Performance of the proposed algorithm is compared against a number of widely used text localization algorithms and the results show that this method can quickly and effectively localize and extract text regions from real scenes and can be used in mobile robot navigation under an indoor environment to detect text based landmarks.

*Index Terms*— Mobile Robot Navigation, text localization, morphological operator, SVM classifier, discrete wavelet transforms.

## I. INTRODUCTION

Automatic Region of Interest (ROI) detection is an active research area in the design of machine vision systems, and is used in many applications, such as autonomous mobile robot navigation, tourist, assistant stems, navigational aids for automobiles, vehicle license plate detection and recognition, etc. Using bottom-up image processing algorithms to predict human eye fixations or mimic human perception and cognition and extract the relevant embedded information in images has been widely applied in automatic ROI detection systems.

Psychophysical and neurophysiologic searchers find that two levels of content always draw human visual attention, namely the low level perceptual content, such as color/intensity, texture, edges, geometric shapes, and high level semantic content such as human faces, text, vehicles, etc [1]. In vision-based mobile robot navigation, many potential landmarks such as nameplates and information signs contain text. By locating those text landmarks, it can provide feature correspondence in a sequence of video frames to reconstruct 3D models. Furthermore, text embedded in images contains large quantities of useful semantic information which can be used to fully understand images. Text recognition is very useful in high level robot navigational tasks, such as path planning and goal-driven navigation. Therefore, scene text is an important feature to be extracted. Many methods have been proposed for extraction text information from images and video frames. However there is no a general method that can detect and find texts in all conditions. [1, 2].Generally, two types of text appear in images[1]:

1) Caption text: which is artificially overlaid on the image (Fig.1)
2) Scene text: which exists naturally in the image (Fig.2).

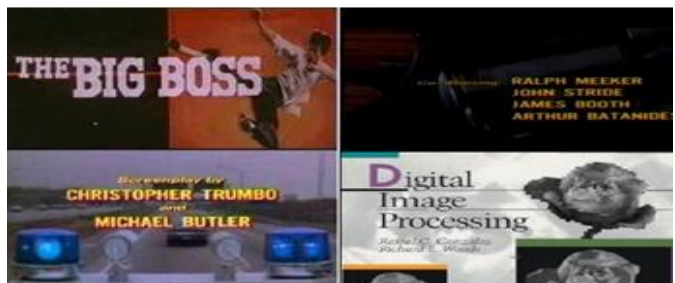
Fig.1 Caption text

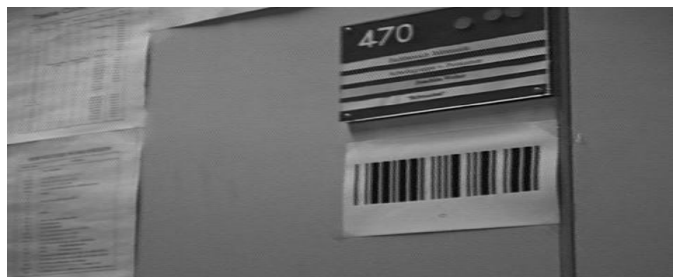
Fig.2 Scene text

As it is shown in Fig.1 and Fig.2, extracting scene text from images can be a difficult task and in Robot Navigation we encounter with this images. Even if we can detect existence of text in image and find regions that these texts are located, then we may feed these regions to an OCR system to extract the text information. For this reason, we need to extract features that can be good descriptor for text existence. Many features that describe existence of text in image are proposed such as texture features [2], color [6], histogram [5] and contrast [6]. In this paper we are interested in features that are accurate, fast and easy to implement.

Many approaches are proposed for text detection and localization in image. Approaches based on region use region color features. These methods are applied in two ways: Methods base on connected component [1] and methods based on edge [2] and a combination of them [3].

In [4] a four-step method based on connected components is proposed for text localization. This method is able to extract text from multiple components under various illumination conditions. But this method has several restrictions related to the text alignment and color.

In another method [5] the color range is quantized by peeks of color histogram. This method is based on an assumption that the text regions are clustered in color range and fill large area of the image. This method is not appropriate for environments with complex background. A method proposed by [1] is based on single color property and use split and merge algorithm to localize text in images, then extract very great or small segments, and do dilation operation and then use motion information and contrast analysis to improve extracted results. In spite of high precision of this method, it is very slow.

The method in [6] uses morphological operators to extract candidate text region and applies a rule based method such as width and length restriction to extract text region with high precision. Although this method has high speed, but for text with complex background has low precision.

The rest of this paper is organized as follows: Section 2 presents morphological operators. Section 3 describes proposed method and finally experimental results and conclusion are presented in section 4 and 5 respectively.

## II. MORPHOLOGICAL OPERATORS

The proposed method uses a new set of morphological operators to extract high contrast regions. We describe briefly morphological operators before explaining proposed method.

In this section a number of morphological operators used for extracting text candidate regions are described. Using of morphological operators has many advantages such as high computation speed, being invariant to several geometrical transformations like translation, rotation, and scaling and it can extract all areas containing text [6, 7, and 8].

Suppose that $S_{m,n}$ is a m×n structure element. Fig.3 shows two such structure elements. Based on $S_{m,n}$, erosion, dilation, differencing, closing and opening operators are defined as following:

Dilation:
$$(I \oplus S)(x) = \max_{\substack{z \in S \\ x-z \in I}} \{I(x-z) + S(z)\} \quad (1)$$

Erosion:
$$(I \ominus S)(x) = \min_{z \in S} \{I(x+z) - S(z)\} \quad (2)$$

Opening:
$$I \circ S = (I \ominus S) \oplus S \quad (3)$$

Closing:
$$I \bullet S = (I \oplus S) \ominus S \quad (4)$$

Where I(x,y) is intensity image. If I1 and I2 are two images, then differencing operator is defined as follows:

$$D(I_1, I_2) = |I_1(x, y) - I_2(x, y)| \quad (5)$$

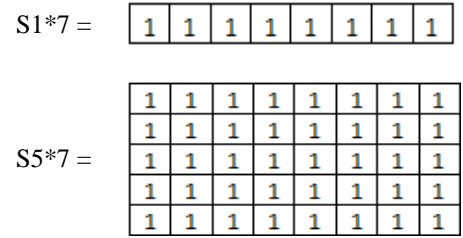

Fig.3) two structure elements

## III. PROPOSED METHOD

As stated before, there are many methods to extract text from images for robot navigation. Because of the needs for a method that can extract text regions with high speed and high precision in real time environments, in this paper a new two step method is proposed to extract location of text in images that meets above requirements. In first step a new set of morphological operators with special sequence applied to extract text candidate regions. After the text candidate regions are extracted, a SVM classifier is used for classifying text candidate regions. In this step texture features that are well illustrators for text existence in images are extracted.

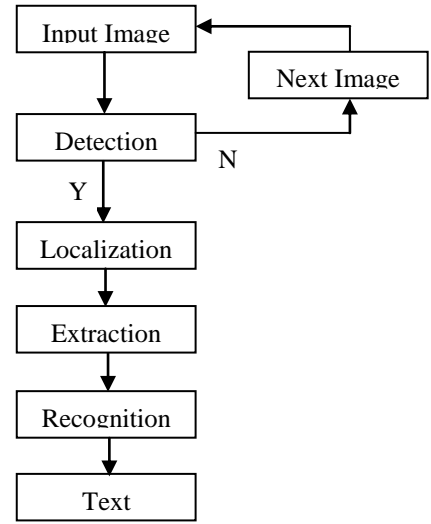

Fig.5) Proposed text extraction system

Fig.5 shows general diagram of the a text extraction system. This system includes following sub systems:
1) Detection: This step is used to detect text existence in images. Input of this step is robot vision images that may

contain text. In the output it produces a binary decision showing the existence of text in image. If image don't have any text, then next image is processed.
2) Localization: After that existence of text in image is verified, text locations have to be extracted. Output of this step is rectangle regions that contain text inside. In this paper we propose new method for this step.
3) Extraction: After that location of text region is found, then the text has to be extracted from background and provided for OCR system.
4) Recognition: This step is final step of the text extraction system. There are many software for do this step, that are called OCR, that recognize text with high accurate.

### A. Extracting text candidate regions

In this section a number of morphological operators used for extracting text candidate regions are described. Using of morphological operators has many advantages such as high computation

Figure 6 shows the proposed algorithm for text extraction. At first a smoothing operator is used to reduce noise with structure element S 3,4. Then opening and closing operators with structure element S_1,5 are computed to create two images A and B at output. For detection of edges, the differencing operators are used on images A and B. For improving the edges, and filling small spaces between characters, closing operator are applied on edge images. Then to create binary image, the thresholding operator is used.

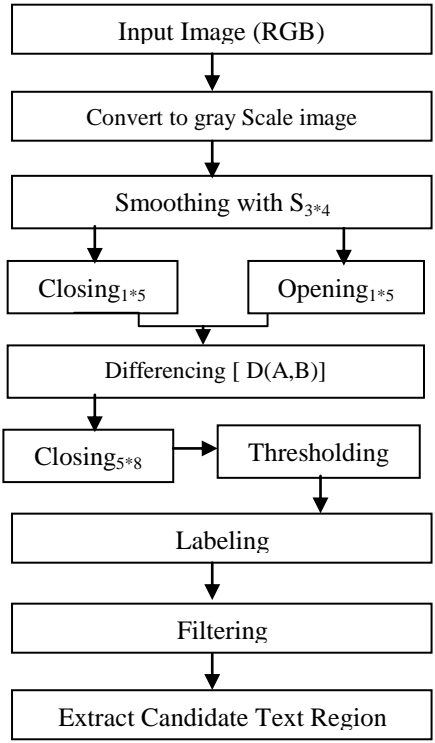

Fig.6) Proposed diagram for extract text candidate regions.

In this method, threshold value is selected between 0.8 and 1.2. After image thresholding, the labeling operation is applied to extract text candidate regions. In this step pixels that are connected to each other, create a connected component which is called text candidate region. As the text in images has some constraints such as length, width and ratio between text length and width, these constraints is used to select some of the extracted text candidate regions to be sent to next step. a candidate region is sent to the next step if it can satisfy following conditions:

$$W_R / H_R > 1.5 \qquad (8)$$

$$density = \frac{Area\ of\ R}{W_R \times H_R} > 0.1 \qquad (9)$$

where R is output binary image from previous step, WR and HR are width and length of text candidate region respectively.

As generally the length of text line is greater than its width, the first rule compares width and length of region. the second role declares that, the density of R should be large enough. Numerator of equation (9), is the number of pixels in text candidate region that have value of "1" in binary image.

As the contrast property is invariant to rotation and high noise tolerance. Therefore this property has important role in detection of text lines. The morphological operators can be computed using convolution operation. There are many high speed algorithms for this operation. Some of them are implemented on DSP and FPGA board, thus the proposed method can be used in real time environment.

### B. Final selection step with use of SVM

In this section a number of morphological operators used for extracting text candidate regions are described. Using of morphological operators has many advantages such as high computation

After that text candidate regions are selected, the features that are appropriate descriptor for text existence are extracted and used by SVM classifier to decide on the existence of text in the region. Output of this step is regions that have been identified as text region by proposed algorithm.

If the extracted features are not appropriate descriptor for text existence, then the SVM may not classify these regions accurately. Therefore, we propose a number of texture features that are extracted in discrete wavelet transform (DWT) domain. Also as DWT can be computed using a convolution operator, thus this step can be applied with very fast speed. Doermann in [11] has proved that the following features in DWT domain are appropriate descriptor for text existence in image. Pay attention to these features will be extracted from the original image. If the text candidate region size has be M × N, the mean and the second and the third central moments of this region are defined as following:

$$m(T) = \frac{1}{M \times N} \sum_{i=0}^{M-1} \sum_{j=0}^{N-1} T(i,j) \qquad (10)$$

$$\mu_2(T) = \frac{1}{M \times N} \sum_{i=0}^{M-1} \sum_{j=0}^{N-1} (T(i,j) - m(T))^2 \qquad (11)$$

$$\mu_3(T) = \frac{1}{M \times N} \sum_{i=0}^{M-1} \sum_{j=0}^{N-1} (T(i,j) - m(T))^3 \qquad (12)$$

Where, $T(i,j)$ is the wavelet coefficient of pixel point $(i,j)$. Then, these features have to be computed for each three wavelet sub band areas including HH, HL and LH. Therefore a set of nine features are extracted for each text candidate region. Afterward we have to collect many images (include text and don't include text) and then extract these features from the text candidate regions. Then we have to train SVM classifier with these images to obtain boundary between true and false samples.

## IV. EXPERIMENTAL RESULT

In this paper we use two data sets to do experiments. First data set images collect by authors. The images are various with font, color, size, direction, background, texture, contrast and quality. Another data set is obtained from [13] that has 45 samples. We use second data set to compare proposed method with some other methods.

Fig.7 shows results of proposed method to find location of texts in image. It's clear that the first step of proposed method has been able to extract all text candidate regions, and then in the second step only candidate regions that have text are extracted. Fig.7 shows that, our method can extract all text location in image with complex background.

To evaluate the proposed method with other methods three criteria are defined as below:

$$\text{Recall} = \frac{Number\ of\ correctly\ \det ected\ text}{Number\ of\ text} \quad (13)$$

$$False\ Alarm\ Rate = \frac{Number\ of\ falsly\ \det ected\ text}{Number\ of\ \det ected\ text} \quad (14)$$

$$Speed = Number\ of\ processed\ images\ per\ \sec ond \quad (15)$$

Finally, we used above criteria and compared our method with a number of methods. We obtained following results:

Table.1) Compare performance of proposed method with 3 methods

| Method | Speed | False Alarm Rate (%) | Recall(%) |
|---|---|---|---|
| Proposed Method | 10.2 | 2.5 | 96.1 |
| Method[3] | 8.3 | 2.4 | 94.2 |
| Method[6] | 8.2 | 5.3 | 95.3 |
| Method[11] | 1.5 | 5.6 | 91.4 |
| Method[12] | 2.2 | 8.1 | 94.3 |

As seen in Table 1, the proposed method can find text areas with 96.1% accuracy that compared to other methods, more text areas are extracted. Also, because of our method can be applied in real time, number of images per second that can be reviewed is more than other methods.

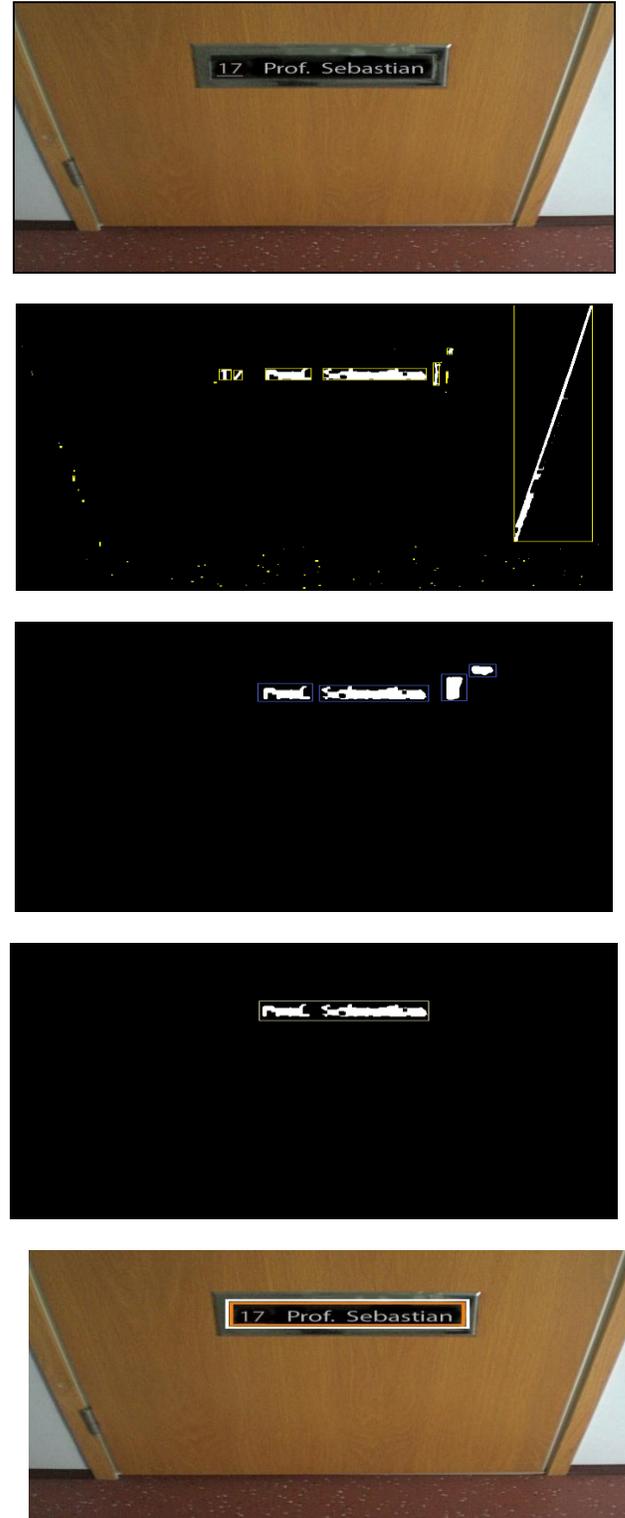

Fig.7) images from up to bottom: original image- text candidate regions are extracted using morphological operators- text candidate region with applied constraint (8), (9) – Final text region are extracted using SVM- Final output on original image

## V. CONCLUSION

In this paper, a novel two-step text localization method for Indoor Mobile Robot Navigation is introduced. The important characteristics of our method are: 1) can be implemented in real time environments. 2) Extract all text regions with high accuracy. In the first step, a new set of morphological operators are proposed to extract candidate text regions. In the second step, a set of nine features are extracted that can be extracted easily and can be computed with high speed and ultimately cause to high precision in classification.

REFERENCE

[1] Sebastian Schneegans." Using RFID Snapshots for Mobile Robot Self-Localization ". roceedings of the 3rd European Conference on Mobile Robots, September 19-21, 2007, pp. 241-246

[2] Sooyong Lee and Jae-Bok Song," Mobile Robot Localization using Range Sensors : Consecutive Scanning and Cooperative Scanning", International Journal of Control, Automation, and Systems, vol. 3, no. 1, pp. 1-14, March 2005

[3] X. Liu and J. Samarabandu, "An edge-based text region extraction algorithm for indoor mobile robot navigation," in Proc. of the IEEE International Conference on Mechatronics and Automation (ICMA 2005)

[4] J. Ohya, A. Shio, S. Akamatsu, "Recognizing characters in scene images", IEEE Trans. Pattern Anal. Mach., Vol 16 (1994), Pages 214–224.

[5] JY. Zhong, K. Karu, A.K. Jain, "Locating text in complex color images", Pattern Recognition vol 28 (1995), pages 1523–1535.

[6] Jui-Chen Wu · Jun-Wei Hsieh · Yung-Sheng Chen."Morphology-based text line extraction". Machine Vision and Applications, vol 19 (2008), pages 195–207

[7] C.R. Giardini, E.R. Dougherty, "Morphological methods in image and signal processsing",Prentice Hall, (1988).

[8] Gonzalez, R.C. and Woods, R.E."Digital image processing", Prentice Hall, (2002)

[9] K Yoshida, A Sakurai , "Machine Learning" , Encyclopedia of Information Systems, (2004), pages 103-114

[10] M.A. Kumar and M. Gopal, "Least squares twin support vector machines for pattern classification", Expert Systems with Applications vol 36 (2009) . pages 7535–7543

[11] H. Li, D. Doermann, O. Kia, "Automatic text detection and tracking in digital video", IEEETransactions on Image Processing, vol 9 (2000) , pages 147–156

[12] R. Lienhart, A. Wernicke, "Localizing and segmenting text in images and videos", IEEE Transactions on Circuits and Systems, vol 12 (2002), pages 256–268

[13] X.S. Hua,W.Y. Liu, H.J. Zhang, "An automatic performance evaluation protocol for video text detection algorithms", IEEE Transactions on Circuits and Systems for Video Technology vol 14 (2004), pages 498–507.

[14] M.D. Wilson, "Support Vector Machines" , Encyclopedia of Ecology, (2008), pages 3431-3437